\documentclass[sigconf]{acmart}
\usepackage{latexsym}

\usepackage{microtype}
\usepackage{graphics}
\usepackage{arydshln}
\usepackage{booktabs}
\usepackage{amsmath}
\usepackage{soul}
\usepackage{diagbox}
\usepackage{graphicx}
\usepackage{float}
\usepackage{enumitem}
\usepackage{multirow}
\usepackage{todonotes}

\usepackage{xcolor,colortbl}
\definecolor{LightGreen}{rgb}{0.8,1,0.89}
\newcommand{\model}{\textsf{ConSum}}
\newcommand{\data}{\textsf{MEMO}}

\copyrightyear{2022}
\acmYear{2022}
\setcopyright{acmcopyright}\acmConference[KDD '22]{Proceedings of the 28th ACM SIGKDD Conference on Knowledge Discovery and Data Mining}{August 14--18, 2022}{Washington, DC, USA}
\acmBooktitle{Proceedings of the 28th ACM SIGKDD Conference on Knowledge Discovery and Data Mining (KDD '22), August 14--18, 2022, Washington, DC, USA}
\acmPrice{15.00}
\acmDOI{10.1145/3534678.3539187}
\acmISBN{978-1-4503-9385-0/22/08}



\begin{document}
\title{Counseling Summarization using Mental Health Knowledge Guided Utterance Filtering}

\author{Aseem Srivastava\textsuperscript{$1$}, Tharun Suresh\textsuperscript{$1$}, Sarah Peregrine (Grin) Lord\textsuperscript{$2,3$}, \\Md. Shad Akhtar\textsuperscript{$1$}, Tanmoy Chakraborty\textsuperscript{$1$} }
\affiliation{
\country{
    {
        \textsuperscript{$1$}IIIT-Delhi, India; \textsuperscript{$2$}University of Washington; \textsuperscript{$3$}Mpathic.ai
    }
}
}

\email{{aseems, tharun20119, shad.akhtar, tanmoy}@iiitd.ac.in;  grin@mpathic.ai}

\renewcommand{\shortauthors}{Aseem Srivastava, Tharun Suresh, Sarah Peregrine (Grin) Lord, Md. Shad Akhtar, Tanmoy Chakraborty}

\begin{abstract}
The psychotherapy intervention technique is a multifaceted conversation between a therapist and a patient. Unlike general clinical discussions, psychotherapy's core components (viz. symptoms) are hard to distinguish, thus becoming a complex problem to summarize later. A structured counseling conversation may contain discussions about symptoms, history of mental health issues, or the discovery of the patient's behavior. It may also contain discussion filler words irrelevant to a clinical summary. We refer to these elements of structured psychotherapy as {\em counseling components}. In this paper, the aim is mental health counseling summarization to build upon domain knowledge and to help clinicians quickly glean meaning. We create a new dataset after annotating $12.9K$ utterances of counseling components and reference summaries for each dialogue. Further, we propose \model, a novel counseling-component guided summarization model. \model\ undergoes three independent modules. First, to assess the presence of depressive symptoms, it filters utterances utilizing the {\em Patient Health Questionnaire} (PHQ-9), while the second and third modules aim to classify counseling components. At last, we propose a {\em problem-specific Mental Health Information Capture} (MHIC) evaluation metric for counseling summaries. Our comparative study shows that we improve on performance and generate cohesive, semantic, and coherent summaries. We comprehensively analyze the generated summaries to investigate the capturing of psychotherapy elements. Human and clinical evaluations on the summary show that \model\ generates quality summary. Further, mental health experts validate the clinical acceptability of the \model. Lastly, we discuss the uniqueness in mental health counseling summarization in the real world and show evidences of its deployment on an online application with the support of \textit{mpathic.ai}.
\end{abstract}

\maketitle

\begin{figure}[t]
  \centering
  \includegraphics[width=\columnwidth]{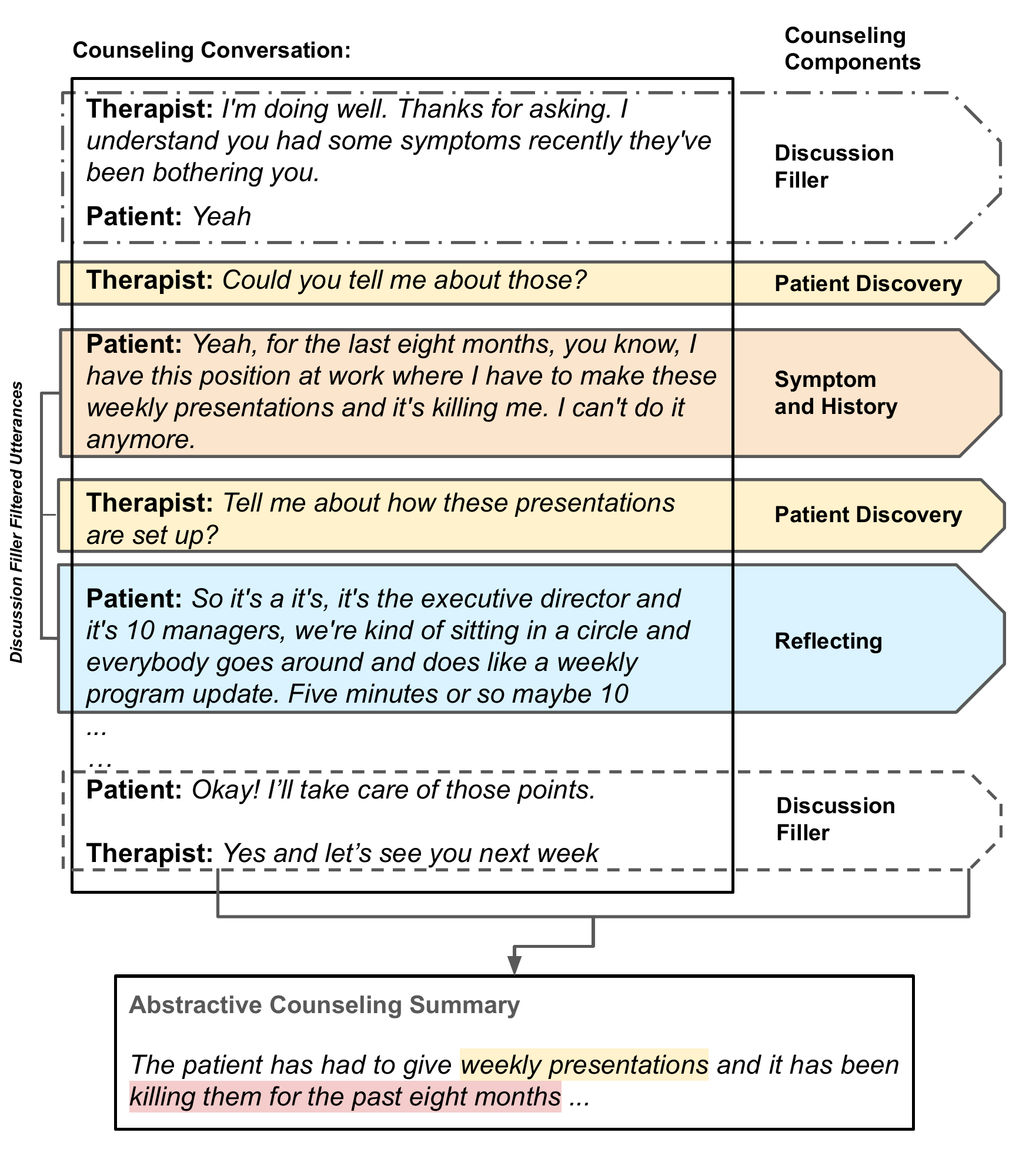}
  \vspace{-10mm}
  \caption{A sample counseling session instance from the \data\  dataset. \textbf{\textit{{\color{orange!60!black}Symptom and History}, {\color{cyan!50!black}Patient Discovery}, {\color{yellow!50!black}Reflecting}, and \textit{Discussion Filler}}} are psychotherapy elements. The summaries pertaining to this truncated snippet of conversation is shown at the bottom. \emph{Note: The summary and conversation are truncated for brevity.}}
  \label{fig:taskfigure}
  \vspace{-5mm}
\end{figure}

\section{Introduction}
Mental health counseling is one of the front-line defenses against mental health illness. In medical and primary care settings, the doctor follows a highly-structured assessment approach that elicits specific information about the patient's medical problems to rule out different diagnoses. In psychotherapy counseling sessions, patients take center stage in elucidating their situation with subtle details. The therapist introduces diverse auxiliary context in conversations to put the patient at ease, discuss events happening in the patient's recent past including the feelings, reflections and emotions that the patient experiences, and other relevant topics. Follow-up conversations with the patient are also vital for a successful treatment. The points of the counseling session that are crucial for continuity of care in follow-up and treatment planning include: (a) the patient's presenting problem, (b) symptoms and diagnosis, (c) treatments (current and prior), (d) mental status and risk assessment, and (e) other varied context and history. An efficient way to present this information to a therapist is to summarize the therapist-patient conversation.  
The process highlighted earlier, differentiates a mental health counseling session from other medical conversations, and extraction of relevant utterances is key to avoid noise in generated summaries. 

In this paper, we aim for the task of counseling conversation summarization. Our work incorporates domain knowledge to generate summaries that inherit essential psychotherapy elements. Earlier studies \cite{see_get_2017,nallapati_abstractive_2016} show attention based architectures to generate summary. Recent studies \cite{afzal_clinical_2020, song_summarizing_2020} use segmentation of the conversation based on topic similarities in summarizing dialogue. Extending the ideas by incorporating mental health domain knowledge, we propose `counseling components' for each utterance to include the understanding associated with counseling conversation. Comprehensive analyses of counseling conversations show that the conversation aims to discuss the reasons for the problem, discover the patient’s insights, and reflect on their past. Hence, we create the dataset, \data\footnote{MEMO: Mental hEalth suMmarizatOn dataset} where we annotate 12.9K utterances with the following four labels - \emph{symptom and history, patient discovery, reflecting, discussion filler}. Also, we annotate and validate the dataset with the help of mental health experts.

The state-of-the-art models are capable of generating semantically rich text; however, to pick the fitting symptoms and essence of the counseling problem is still a challenge. In this paper, we incorporated mental health domain knowledge using the Patient Health Questionnaire (PHQ-9) \citep{kroenke_phq-9_2001} to represent an established set of questions to assess the patient’s mental health symptoms. The PHQ-9 specifically assesses depressive symptoms but represents an example of symptom-specific domain knowledge. This approach could be generalized to other frequently-used assessments focusing on other symptom profiles like anxiety (e.g., GAD-7). Taking advantage of domain knowledge, we identify utterances categorized as discussion filler or irrelevant and will not be prioritized. Labeling across different counseling conversations aids the model to map similarities and serve as the supervision signal to attend to salient utterances and extract appropriate information to be a part of the summaries.

We propose \model\footnote{ConSum: \textbf{Co}u\textbf{n}seling \textbf{Sum}marization}, a mental health counseling summarization architecture. The complete pipeline of the model  works in three different modules: (a) The first module (MH-Know) exploits domain knowledge to filter utterances irrelevant to the counseling summarization task. (b) Secondly, we propose a discussion filler classification module that filters irrelevant utterances like \textit{`yeah', `ummm'}, etc. (c) The third module predicts the counseling components to utilize the structured therapy knowledge for summarization task.  

We run our experiments on the \data\ dataset. Evaluations show improvements across all baselines and discuss the effect of counseling components and effective use of domain knowledge in our  \model\ model. Finally, we discuss the uniqueness of counseling summarization and explore the generalizability of \model. Significant contributions of our work are as follows:
\begin{itemize}[leftmargin=*]

    \item We propose \model, a summarization model that exploits mental health domain knowledge and counseling components.
    
    \item We propose a novel counseling summarization dataset - \data\ and present a novel annotation scheme for psychotherapy elements viz. symptom and history, patient discovery, reflecting aspects, and discussion filler in utterances of counseling dialogue.
    
    \item We propose a new problem specific metric to evaluate summaries i.e., \textbf{M}ental \textbf{H}ealth \textbf{I}nformation \textbf{C}apture \textbf{(MHIC)} metric which reasonably evaluates summaries that are most useful from a counseling's perspective.
    
    \item Mental health experts in \textit{mpathic.ai} meticulously analyzed \data. They studied the results on various clinical and linguistic parameters to evaluate the acceptability of \model's performance for commercial and clinical application at scale.

\end{itemize}

\section{Related Work}
\label{section:lit}
Given the critical nature of the medical domain, the pace of adoption of modern deep learning models did not gain enough traction for impactful change. Instead, it was used to address mundane medical procedures. On the other hand, summarization is a long-studied problem in text processing. Earlier, most of the focus and improvements were on extractive summarization, with a drastic shift to abstractive summarization recently. Some of the work done in the medical health domain is discussed below, followed by a section on technical progress in summarization methods.

\subsection{Mental Health and NLP} One of the early works by \citet{chen_query-based_2006} explored a simple information retrieval system to fetch information from medical documents using document summaries. They employed medical ontology to help fine-tune the user query; however, simple summarization methods limited its effectiveness. Later, an NLP-based framework was introduced by \citet{konovalov_biomedical_2010} to identify a personnel's physical exposure (in a battle) and their emotional reaction to it. Their work highlighted the importance of subtle indicators of mental illness and the advantages of diagnosing ``early components" in counseling. \citet{strauss_machine_2013} analyzed clinical forms in mental health to cut down on the arduous task of manual analysis using machine learning. Their work paved the way to explore machine learning as a tool to automate tasks dependably, to a certain extent.\par
Secondly, certain efforts helped address mental health problems specifically. \citet{kennedy_canadian_2016} showed a comprehensive study about the management of adults with Major Depressive Disorder (MDD). Their analysis identifies different stages of depression and discusses an appropriate level of antidepressant to choose based on the patient's age, anxiety levels, and long episode duration. Subsequently, the prescribed antidepressants are monitored for satisfactory patient response. This notion motivates the identification of key indicators of the patient's conversation with the therapist and follow-up sessions that rely on the patient's past interactions (and prescriptions). \citet{tran_predicting_2017} experimented with neural network models to predict the mental illness condition from neuropsychiatric notes. These notes comprised 300 words on average about the patient's present illness and events associated with it, followed by a psychiatric review system that mentions mental illness associated with the patient. Despite using various models such as RNN, BiRNN, and LSTM, it is a straightforward classification problem among the 13 predefined mental illnesses.\par
In deep learning based summarization models, extractive summarization \cite{chen_modified_2020}  addressed the problem of doctors going through elaborate discharge diagnoses by summarizing them using the BERT model. They used character-level tokens to reduce the parameter size of the BERT model to deploy in low-resource setups. However, there were no sizeable efforts made in explicitly capturing medical information in summaries.\par 
Some of the contemporary approaches include the work by \citet{ive_generation_2020} where they generated artificial datasets from genuine mental health records. Besides, \citet{afzal_clinical_2020} focused on identifying PICO (Patient/Problem, Intervention, Comparison, and Outcome) on a well-structured sequence of sentences from medical documents to present a concise extractive summary only. Recently, to filter relevant utterances, \citet{info:doi/10.2196/20865} used domain knowledge from patient health care questionnaire (PHQ-9)  to build knowledge graphs. They used an unsupervised approach to build abstractive summaries for the counseling session. \citet{zhang-etal-2021-leveraging-pretrained} showed the generation of summary in two steps, one for building chunks of partial summaries and then second for fusing those chunks to generate final summary. Moreover, \citet{9508585} showed an online application developed based on medical domain-specific annotation and information extraction.  

\vspace{-2mm}
\subsection{Dialogue Summarization} 
Broadly, the two categories of dialogue summarization tasks are extractive (utterance filtering) and abstractive (semantically rich). Summarization has been long studied under various settings \citep{ATRI2021107152, dey-etal-2020-corpora}. Earlier, \citet{chen_spoken_2011} proposed a graph-based approach for extractive summarization wherein topical similarities between each sentence are identified using Probabilistic Latent Semantic Analysis (PLSA). \citet{saggion_automatic_2012} highlighted two broad extractive summarization techniques, which are \emph{superficial techniques} including statistical-based approach and \emph{knowledge-based} methods for pre-training from a large corpora. \par
 An observation on conversation summaries reveals that capturing salient points with a coherent narration (abstractive) is the most succinct format to express. Hence, abstractive summarization gained traction with early work by \citet{nallapati_abstractive_2016} exploring encoder-decoder architecture which was further extended with suitable attention mechanism by \citet{vaswani_attention_2017} for abstractive summarization. Then, a mix of extractive and abstractive summarization with a suitable copy model was introduced by \citet{see_get_2017}. They used a generator model for abstractive summarization, a copy model for extractive summarization, and a coverage vector to limit repetition in summarization. In the sequence, \citet{chen_fast_2018} proposed a reinforcement learning-based approach to extract the salient sentences and concised them to summary sentences with an abstractive network. They employed a specific emphasis to reduce redundancy in the extracted utterances from the conversation. \par
 Further nuanced observation of the relation between salient sentence extraction and conversation utterances shows the dependence on certain groups of utterances. On this basis, recent work by \citet{narayan_dont_2018} introduced topic distribution based on Latent Dirichlet Allocation \cite{blei_latent_2003}. Extending on the idea of topic segregation, \citet{song_summarizing_2020} used three labels to tag each utterance with - Problem Description, Diagnosis, and Other. They showed that adding utterance labels aids in summarization performance. Nevertheless, since their reference summaries are extractive, the complexities of their models are limited. 
Further, \citet{zhang_pegasus_2020} and \citet{rael_exploring_nodate} pushed the boundaries of abstractive summarization by generating semantically rich sentences, diversely adapted to different generative tasks. In the clinical conversation domain,  \citet{doi:10.1177/1460458220951719} and \citet{krishna-etal-2021-generating} followed the approach of important utterance selection for medical conversation summarization. 

\section{\data: Mental Health Summarization Dataset}
\label{section:dataset}

We create a novel dataset for the task of counseling conversation summarization in which we extend data collected from the publicly available counseling conversation dataset -- HOPE \cite{malhotra2021speaker}. The dataset contains $12.9K$ utterances from $212$ counseling conversations between therapist and patient. The dataset is gathered from multiple counseling videos on publicly available platforms like YouTube. It belongs to diverse demographic groups with distinct mental health matters and several therapists, helping researchers propose a generalized approach. Further, the authors extracted transcriptions from the videos and pre-processed them. Collected dialogues are dyadic, in which patients and therapists are the only interlocutors. Since our task aims for counseling summarization built upon domain knowledge, we extend the HOPE dataset to annotate psychotherapy elements and counseling summary. We consulted with a team of the leading mental health experts and proposed a hierarchical labeling structure for each utterance. We discuss the details related to annotation in Section \ref{sec:annotation_process}. The same experts helped us develop annotation guidelines and validate the annotations. We now discuss the extended dataset, which we call \data.

\begin{figure}[!t]
    \centering
    \includegraphics[width=0.5\textwidth]{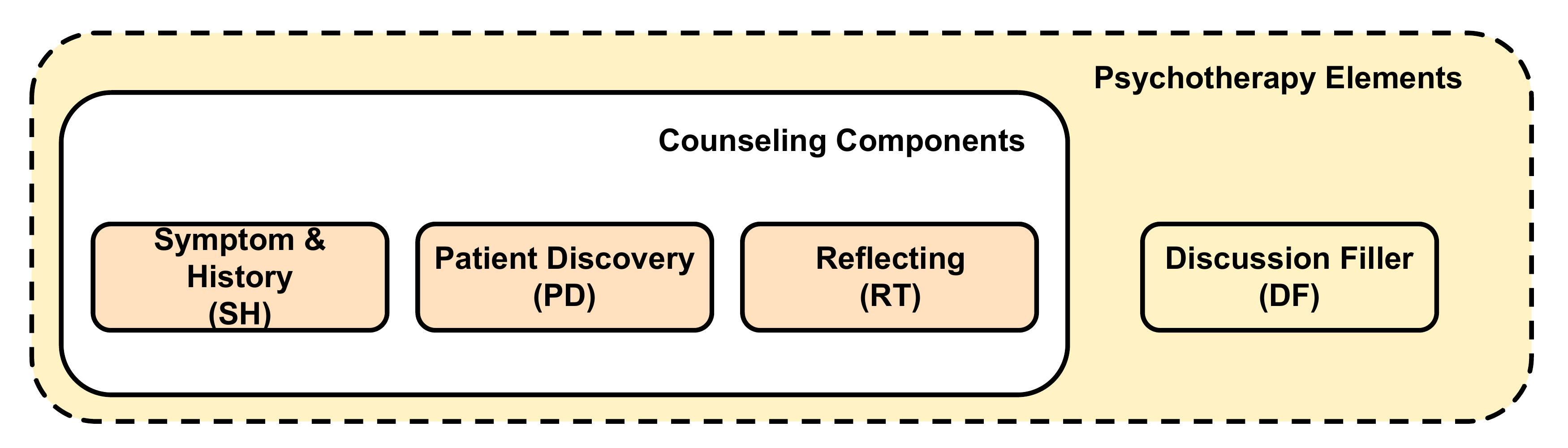}
    \caption{Hierarchical labeling for each utterance. Counseling components including \emph{SH, PD} and {\em RT} essentially contribute in summary generation. }
    \label{fig:annotations}
    \vspace{-5mm}
\end{figure}

\label{sec:memo}
Next, the task is to add psychotherapy elements to these dyadic conversations. We observe that the conversations can contain three essential counseling components, namely \textit{symptom and history, patient discovery}, and {\em reflecting} utterances as shown in Figure \ref{fig:annotations}. The rest of the conversation between therapist and patient falls under the category of \textit{discussion filler}. Furthermore, to ensure the reference summaries represent rich knowledge considering the clinical and linguistic perspectives and contained psychotherapy elements, we follow the annotation guidelines designed by the team of mental health experts.

\subsection{Annotation Process}
\label{sec:annotation_process}
Counseling conversations are typically challenging as the patients are reluctant to express themselves in front of therapists. As a consequence, therapist articulately involves the patient into discussion. In support of this, we meticulously analyze counseling conversations to symptom and history of the mental health problem, the discovery of the patient’s behavior, and insights into the past story reflecting on the patient’s current situation. The rest of the counseling conversation is generally enclosed with discussion filler phrases. To better understand the counseling and support summarization task, focusing only on a subset of knowledge-enhanced utterances, we annotate counseling components and discussion filler for all utterances.
With the help of a comprehensive annotation process in place, the techniques used by therapists are highlighted with each utterance. Therapists can be significantly helped with AI based tools adhering to such domain-specific annotation guidelines\footnote{https://familytherapybasics.com/blog/therapy-case-summary}. With the help of guidelines, the annotations keep track of the therapy techniques; therefore, the summaries generated are succinct in their information. We discuss more about annotation process of psychotherapy elements in Appendix \ref{annotation_appendix}.

\textbf{Psychotherapy elements}: Mental health therapy sessions are composed of counseling components that effectively contribute towards successful interventions and discussion filler which barely add any relevance to the summary generation. We labeled utterances with four fine-grained labels:

\begin{itemize}[leftmargin=*]
    \item \underline{Symptom and History (SH)}: It captures the utterances that comprise the most insightful information for the therapist to assess the patient's situation. A few examples are mentioned below:
    
\begin{table}[H]
    \centering
        \scalebox{0.97}{
        \begin{tabular}{l|p{6 cm}}
        Speaker & Examples \\  \hline
        Patient & \emph{``Ever since I moved into the apartment I have been feeling anxious"} \\ \hline
        Patient & \emph{'This has been the first time I have experienced mood swings"} \\ \hline
        Patient & \emph{``I would be so stressed afterward. And worried I would make another mistake"} \\ \hline
    \end{tabular}}
    \label{tab:sr_examples}
\end{table}

    \item \underline{Patient Discovery (PD)}: The patients coming in for a counseling session arrive with complex thoughts. The therapist tries to build therapeutic relationships to calm patients to unravel their thoughts. Examples are as follows:

\begin{table}[H]
    \centering
        \scalebox{0.97}{
        \begin{tabular}{l|p{6 cm}}
        Speaker & Examples \\  \hline
        Therapist & \emph{``Don't worry if it lasts for several days"} \\ \hline
        Therapist & \emph{``I believe sleep is not an issue for you anymore now that you are getting eight hours of sleep everyday"} \\ \hline
        Therapist & \emph{``Any drug or alcohol use?"}, \emph{``Has it ever happened to you?"} \\\hline
    \end{tabular}}
    \label{tab:sr_examples}
\end{table}

    \item  \underline{Reflecting (RT)}: Therapist utterances are concise most of the time, which is to ensure there is sufficient space for the patient to express themselves. Patients are encouraged to share their stories and events of their lives. On the other hand, the therapist understands the patient by giving them an imaginary situation to assess their actions. Some utterances are shown below:

\begin{table}[H]
    \centering
        \scalebox{0.97}{
        \begin{tabular}{l|p{6 cm}}
        Speaker & Examples \\  \hline
        Therapist & \emph{``It's been bothering you for a few weeks, something new, something a little frightening for you \dots"}\\ \hline
        Patient & \emph{``By the time it it's my turn to present I start. I just, I'm almost like just \dots"} \\ \hline
    \end{tabular}}
    \label{tab:sr_examples}
\end{table}

 \item  \underline{Discussion Filler (DF)}: When a therapist and a patient engage in conversation, they articulate some utterances that are peripheral to the session. In this work, we tag them as discussion filler. These include \textit{pleasantries} (`Good morning!'), \textit{non-lexical} fillers (`Ummm'), \textit{acknowledgments} (`Right'), and \textit{restatement of affirmations in subsequent utterances} (`Yeah. Yeah'), etc. Discussion Filler utterances carry little to no relevance in the summary generation.
 
\end{itemize}

\subsection{Data Analysis}
Table \ref{tab:counts} shows data analysis on \data. The dataset consists of an almost equal number of patient and therapist utterances. Therapist utterances are tagged more with discussion filler labels showing that therapists converse to put the patients at ease and agree with them. PD is the most prominent label in the dataset. Patients tend to talk about several auxiliary topics, not their mental health issue, when asked to share their experiences, whereas RT is the least tagged label.

\subsection{Ethical Considerations}
Three domain experts reviewed the annotations at {\em mpathic.ai}, a seed-stage corporation
specializing in conversation analysis in commercial applications, including evaluation of
psychotherapy, coaching, and customer services. All experts are independently licensed
in clinical practice and have backgrounds in automating fidelity of therapy at scale. They
included a board-certified, licensed psychologist (SPL) with over 15 years of experience in
applying machine learning to psychotherapy evaluation, a licensed clinical social worker
specializing in the evaluation of human labeling for machine learning, and doctoral-level
counselor, director of Clinical AI at {\em mpathic.ai}.

Following Sekhon et al.’s framework \cite{Sekhon2017}, they reviewed the summaries for clinical acceptability in
the areas of affective attitude, burden (i.e., the cognitive load of interpreting the summaries),
ethicality, coherence (i.e., how well the summaries were understood), opportunity costs (i.e.,
pros and cons of utilizing the summaries), and perceived effectiveness (i.e., how well they might
perform in a clinical setting). The acceptability parameters were independently evaluated
 and given a ranking of 0-2 on each parameter with 0 being not acceptable, 1,
being acceptable with modification, and 2 being acceptable for commercialization. After
independent ranking, discrepancies were discussed, and an average gold-standard acceptability
the ranking was determined for each summary. Additionally, the experts provided feedback for
commercialization considerations in applications like electronic health care records and quality
assurance.

\section{Proposed Methodology}
In this section, we describe our proposed method, called \model. It encodes dialogue’s utterances and uses three modules in the pipeline to extract complementary information. These three modules operate independently on the selection and filtering of utterances. We use mental health's Patient Health Questionnaire (PHQ-9) \citep{kroenke_phq-9_2001} knowledge similarity (MH-Know) and discussion filler classification to filter utterances from each counseling dialogue. Further, we classify utterances from dialogue for counseling component labels and use these label information to generate abstractive summaries. Figure \ref{fig:modefigure} shows a high-level architecture of \model.

 Consider a dialogue containing $n$ utterances,  ${D =\langle u_1, u_2, \dots u_n \rangle }$. We fine-tune DistilBERT \citep{sanh_distilbert_2020} embeddings to create utterance representation of $d$ dimension. With knowledge-infused utterance selection (MH-Know) module and discussion filler classifier, we generate binary mask arrays from each module, which is fused to filter the utterances. On the other hand, the classification from counseling components classifier augments each utterance with mental health aspects. We describe each module below.

\begin{table}[t]
    \centering
    \caption{Distribution of psychotherapy elements  in \data\ dataset. Here, \textbf{\#D:} Number of dialogues; \textbf{\#U:} Number of utterances; \textbf{U/D:} Utterances per dialogue; \textbf{Sp:} speaker; \textbf{Th:} Therapist; and \textbf{Pt:}. Patient. The train, test, and validation splits are 70:20:10.}
    \label{tab:counts}
    \vspace{-3mm}
    \resizebox{1\columnwidth}{!}
    {
    \begin{tabular}{c c c c c c c c c c c c c c c c}
    \toprule[1pt]
         &  &  &  &  & \multicolumn{6}{c}{\bf Counseling Components} & \\ 
         \cline{6-11}
         &  &  & \multicolumn{2}{c}{\bf Discussion Filler} & \multicolumn{2}{c}{\bf SH} & \multicolumn{2}{c}{\bf RT} & \multicolumn{2}{c}{\bf PD}\\ 
         \cline{4-11}
        \textbf{Split} & \textbf{\#D} & \textbf{Sp.} & U/D & \#U & U/D & \#U & U/D & \#U & U/D & \#U & \textbf{Total} \\
        \toprule[1pt]
        \multirow{2}{*}{\textbf{Train}} & \multirow{2}{*}{152} & Pt & $5.02$ & $764$ & $5.67$ & $862$ & $2.52$ & $383$ & $18.84$ & $2863$ & $4766$\\
         & & Th & $8.10$ & $1232$ & $8.17$ & $1243$ & $4.22$ & $642$ & $10.80$ & $1643$ & $4877$\\
        \cdashline{1-12} 
        \multirow{2}{*}{\textbf{Test}} & \multirow{2}{*}{39} & Pt & $3.43$ & $134$ & $2.95$ & $115$ & $1.00$ & $39$ & $18.38$ & $717$ & $1004$\\
         & & Th & $5.46$ & $213$ & $4.51$ & $176$ & $4.43$ & $173$ & $11.28$ & $440$ & $1006$\\
        \cdashline{1-12}
        \multirow{2}{*}{\textbf{Val}} & \multirow{2}{*}{21} & Pt & $8.09$ & $170$ & $4.23$ & $89$ & $2.28$ & $48$ & $13.80$ & $290$ & $594$\\
         & & Th & $10.48$ & $220$ & $5.57$ & $117$ & $4.66$ & $98$ & $7.38$ & $155$ & $597$\\ \cline{1-12}
         \multirow{2}{*}{\textbf{Total}} & \multirow{2}{*}{212} & Pt & $5.51$ & $1068$ & $4.28$ & $1066$ & $1.60$ & $470$ & $17.00$ & $3870$ & $6364$\\
         & & Th & $8.01$ & $1665$ & $6.08$ & $1536$ & $4.44$ & $913$ & $9.82$ & $2238$ & $6480$\\ 
         \toprule[1pt]

    \end{tabular}}

    \vspace{-5mm}
\end{table}

\begin{itemize}[leftmargin=*]
    \item \textbf{Discussion Filler Classifier (DFC):}
Discussion filler classification is a binary classification task. We use a feed-forward network with $2$ hidden layers to classify each utterance $u_i$ into `counseling’ or `discussion filler’ from the obtained input representations. It compresses representations to $100$ dimensional hidden representation and then to $2$ at the output layer. A dropout of $30\%$ is applied between the linear layers to regularize the model’s performance. The model is trained with adam optimizer to minimize the cross-entropy loss. The output is a mask array labeled $\tau$, where $\tau_i$  represents mask array from dialogue $D_i$. Using independent networks to classify the discussion filler label and the counseling components helps smooth errors better than having a single unified classifier.

\item \textbf{Mental Health Knowledge Infused Utterance Selection (MH-Know):}
Earlier efforts in mental health counseling summarization paid less attention to the domain knowledge. Patient Health Questionnaire (PHQ-9) represents an established set of questionings to assess the patient's mental health condition \cite{kroenke_phq-9_2001}. We use PHQ-9 lexicons \cite{PMID:29707701} to compute similarity between nine questions and input utterance. The intuition is to obtain the most relevant utterances corresponding to knowledge infused questions. We use BERTScore \cite{zhang2020bertscore} to compute the similarity between the utterances $u_i$ and the PHQ-9 lexicons $PHQ = \langle phq_1,phq_2, \dots phq_9 \rangle$. The nine similarity scores, $s = \langle s_1,s_2, \dots s_9 \rangle$ corresponding to each utterance ranges between $0$ to $1$ each. The summation of similarity score denoted by $\psi$ is compared with the hyperparameter tuned value of threshold denoted by $\phi$.
\[
    s_i = bertscore(u_i,phq_i) \Rightarrow \psi_i = \sum_{m=1}^{m=9} s_m
\]

The value of $\phi$ is experimentally chosen to retain maximum domain knowledge and intervention components without compromising other pipelines’ impact. This creates a mask-array,  $\sigma_i = \langle \sigma_1, \sigma_2 \dots \sigma_n \rangle$ containing $1$ for cases where intervention similarity score is less than the threshold and $0$ otherwise. Here, $i$ in $\sigma_i$ represents $i\textsuperscript{th}$ utterance.
\[
  \sigma_i = \begin{cases}
    1, & \text{if $\psi_i \leq \phi$}.\\
    0, & \text{otherwise}
  \end{cases}
\]

\model\ prioritizes the utterances fetched from the MH-Know module along with corresponding counseling tags to generate summary. During experimentation, we use the threshold value of $6$ to extract utterances relevant to the task.

\begin{figure}[t]
  \centering
  \includegraphics[width=\columnwidth]{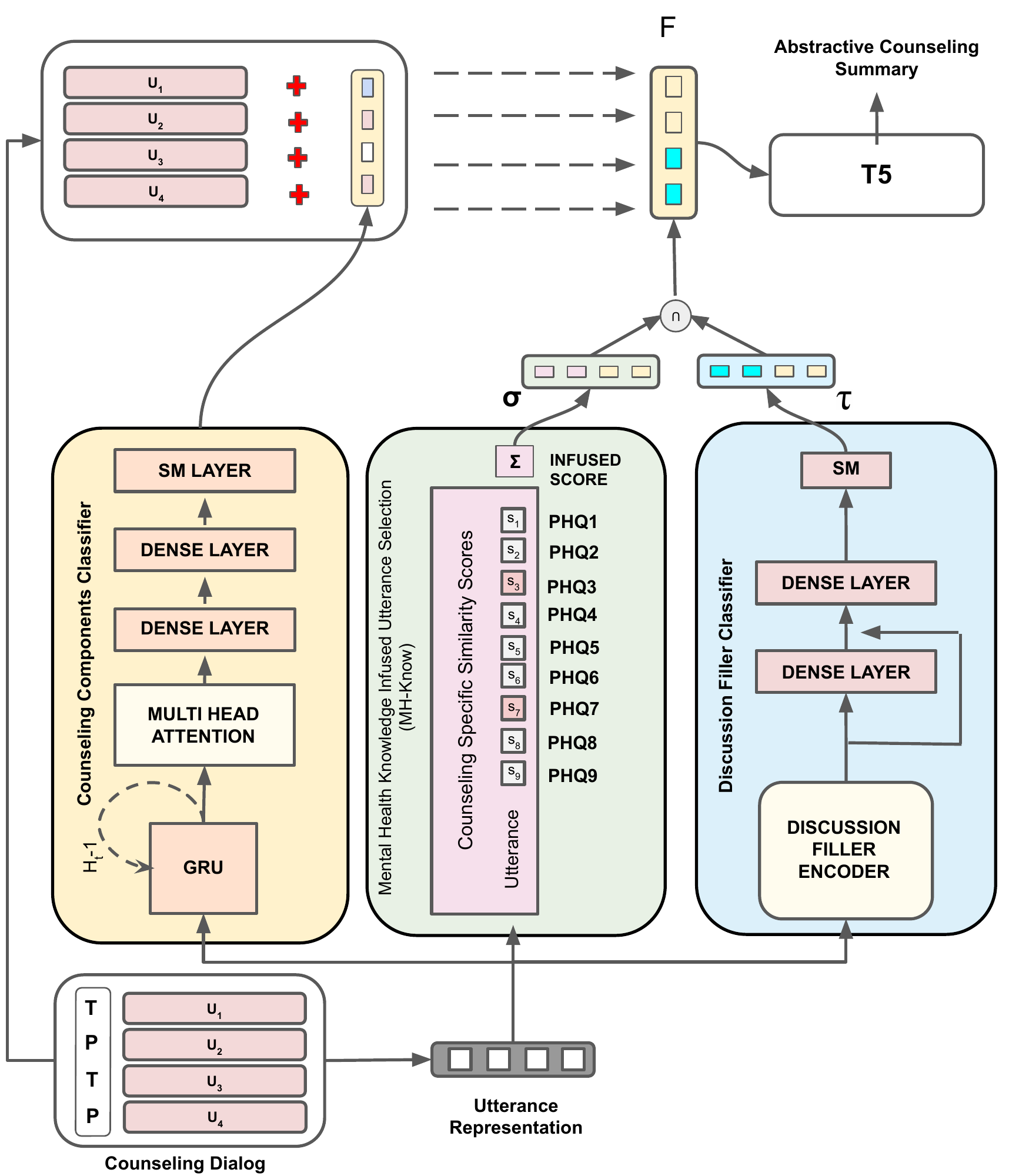}
  \vspace{-6mm}
  \caption{ Model diagram of \model. The Utterance embeddings are reused across three modules in the pipeline.}
  \label{fig:modefigure}
  \vspace{-5mm}
\end{figure}

\item \textbf{Counseling Components Classifier (CCC): }
\label{section:counseling_label_model}
As discussed earlier, the four components are SH, RT, PD, and DF (c.f. Section \ref{sec:annotation_process}). Considering SH, RT and PD as the key components contributing to the summary generation, we frame the problem as a four-class classification task. Hence, for each utterance representation of dimension $d$, we first fetch the \textit{`counseling context'} across utterances. As  the  counseling  progresses,  we  maintain  the context of the dialogue through a GRU layer on top of the utterance representations.
Further, we apply multi-head self-attention following \citet{vaswani_attention_2017}. Representations fused with attention weights are passed through two dense layers to learn hidden representations from $100$ dimensional and then further to $3$. We apply a softmax classifier to predict probabilities corresponding to each class label.

Identification of discussion filler is a domain-independent binary classification task. Therefore, rather than following an end-to-end model for psychotherapy element classification, we opt for a hierarchical model to predict discussion filler and counseling component. In our case, the standard model performs well for DFC. However, we sense the requirement of a model that can learn contextual counseling labels in the summary generation process. The remaining section shows the direct use of counseling labels in the summary generation module. 

\item \textbf{Summary decoder:}
We obtain two mask-arrays from the MH-Know module and DFC. Discussion Filler mask $\tau$ and knowledge-infused mask $\sigma$ are merged as follows. $ F$ denotes the resulting mask array. Further, utterances $U_i$ from the dialogue $D_i$ are concatenated with predicted counseling components (c.f. Section \ref{section:counseling_label_model}) and filtered using the final mask array $F$. The final subset of utterances with domain knowledge and counseling components ($CL_i$) are concatenated, resulting in $G$. It is used as an input to fine-tune the pre-trained T5 model \citep{rael_exploring_nodate} for abstractive summary generation.
$$
F = ( \sigma \cap \tau )' ; G = (U_i \oplus CL_i) \otimes F
$$
\end{itemize}

\section{Experiments and Results}
This section reports our experimental results, comparative study, and other analyses. To evaluate the summaries, we use ROUGE \citep{lin_rouge_nodate} metrics, namely Rouge-1 (R-1), Rouge-2 (R-2), and Rouge-L (R-L). In addition, we show QuestEval (QAE) score and Bleurt Score (BS) to capture the contextual advantage in abstractive summarization. {QuestEval} \citep{scialom_questeval_2021} comprehensively judges the reference and predictions on four aspects -- consistency, coherence, fluency, and relevance by generating question-answer pairs on the given source document, reference, and generated summaries. {Bleurt} \citep{sellam_bleurt_2020} utilizes pre-training on the BERT model using unsupervised techniques with millions of synthetic examples to generalize all possible data distributions. At last, we show human and clinical evaluations to validate the quality of generated summaries. We show hyper-parameters and other experiment details Appendix \ref{app:implementation}. 













\begin{table}[!t]\centering
\caption{Results obtained on \data. We report  Rouge-1 (R-1), Rouge-2 (R-2), Rouge-L (R-L), Bleurt Score (BS) and QuestEval Score (QAE).}\label{tab:baselines}
\vspace{-3mm}
\begin{tabular}{l|rrrrr}\toprule
Model &R-1 &R-2 &R-L &QAE &BS \\
\cmidrule{1-6}

PLM  &34.24 & 11.19 &\textbf{33.35} & 24.34 & -0.8678\\
RankAE  & 25.57 & 3.43 & 24.16 & 29.98 & -1.063\\
SM  & 20.46 & 3.80 & 18.87 & 20.22 & -0.9454 \\
Pegasus & 29.71 & 7.77 & 27.57 &\textbf{36.80} & -0.6130\\
T5 &31.44 &5.63 & 27.38 &33.55 & -0.5655\\
\cdashline{1-6}
\textbf{\model} &\textbf{45.36} &\textbf{15.71} & 24.75 & 25.42 &\textbf{0.3407} \\

\bottomrule
\end{tabular}
\vspace{-3mm}
\end{table}

\subsection{Baselines}
We choose following systems as our baselines - \textbf{Pretrained Language Model (PLM):}  \citet{feng-etal-2021-language} used DialoGPT to segment topics and then generated summaries with BART. \textbf{RankAE:} \citet{Zou_Lin_Zhao_Kang_Jiang_Sun_Zhang_Huang_Liu_2021} used BERT to group utterances into segments and further generated summaries using denoising auto-encoder \citep{vincent_extracting_2008}. \textbf{Segmented Modeling (SM):} \citet{goo_abstractive_2018} leveraged dialogue act information on standard BiLSTM \citep{nallapati_abstractive_2016} architecture. \textbf{Transfer Learning with unified Text-to-Text Transformer (T5):}  \citet{rael_exploring_nodate} used a shared framework on transformer architecture to pre-train on huge custom crawled corpus (C4). \textbf{Pegasus:} \citet{zhang_pegasus_2020} introduced a novel pre-training objective of using gap-sentences-generation and then generated summaries.

\subsection{Results and Ablation Study}
The experimental results on \data\ are shown in Table \ref{tab:baselines}. \model\ performs better than all other baselines with the highest R-1 score of 
$45.36$ and R-2 score of $15.71$ -- significant improvements of $11.12$ and $4.52$ points over the best baseline, respectively. The PLM model turns out to be the second ranked model in both cases with an R-1 score of $34.24$ and an R-2 score of $11.19$. It shows the importance of topic segmentation in a conversational dataset, which aids in the summarization outputs. In contrast, PLM scores the highest R-L score of $33.35$; however, its summaries have a higher structural similarity instead of contextual similarity between its output and the reference summaries. On the other hand, \model\ produces a dominant performance on Bleurt score with a difference of $0.90$ points over the best baseline ($-0.5655$ of T5 vs $0.3407$ of \model). This score indicates contextually superior performance concerning the reference summary.

\begin{table}[!t]\centering
\caption{Ablation study on the effect of different modules of \model. SH, PD, DF and RT are psychotherapy elements. MH-Know and CCC are knowledge infused module and counseling component classifier module respectively. We present our analysis on five scores to understand all aspects of summary generation.}
\label{tab:ablation_result}
\vspace{-3mm}
\resizebox{\columnwidth}{!}{
\begin{tabular}{l|rrrrr}\toprule


Counselling Label &R-1 &R-2 &R-L &QAE &BS \\
\cmidrule{1-6}

\model\ $-$ SH $-$ PD & $20.92$ & $5.00$ & $7.44$ & $20.71$ & $0.0019$ \\

\model\ $-$ PD $-$ RT & $36.00$ & $9.00$ & $9.14$ & $20.47$ & $0.2032$ \\

\model\ $-$ RT $-$ SH & $28.63$ &$8.06$ & $9.55$ & $23.02$ & $-0.0209$ \\

\model\ $-$ SH & $39.77$ & $9.55$ & $8.98$ & $24.11$ & $0.1908$ \\

\model\ $-$ PD & $36.87$ & $10.02$ & $11.22$ & \bf 33.38 & $0.2420$ \\

\model\ $-$ RT & $42.01$ &$9.83$ & $16.50$ & $18.03$ & $0.2060$ \\

\model\ $-$ MH-Know $-$ CCC & $39.67$ &$9.95$ & $12.69$ & $21.19$ & $0.2003$ \\

\model\ $-$ MH-Know & $40.42$ &$10.09$ & $11.00$ & $23.97$ & $0.2429$ \\
\cdashline{1-6}
\model\ & \bf 45.36 & \bf 15.71 & \bf 24.75 & 25.42 & \bf 0.3407 \\




\bottomrule
\end{tabular}}
\vspace{-7mm}
\end{table}

\begin{table*}
\centering
\caption{Comparative analysis on generated summaries between T5 and \model\ model. {\color{green!50!black}Green} denotes primary medical condition information, {\color{cyan!50!black}Cyan} denotes SH information, {\color{brown!60!black}{Brown}} denotes RT information and {\color{red!70!black}Red} denotes errors/irrelevance in generated summaries. Comparison shows $\model$ captures better contextual information compared to T5.}
\label{tab:con_sum_features}
\vspace{-3mm}
\resizebox{\textwidth}{!}{%
\begin{tabular}{c|p{58em}} \toprule
\textbf{Model} & {\centering \textbf{Summaries}} \\\cmidrule{1-2}
T5 & {\em Patient is asked a series of questions to test their housing. Patients {\color{green!50!black}do not want to take drugs}, police arrest them and others in their area. {\color{brown!60!black}They secretly talk about the incident}. The patient lives with friends from different places including the city, where they are arrested or taken into custody. The patients feel they have no medical condition such as diabetes or stroke. {\color{brown!60!black}These occured recently}. The patient has had problems with housing due {\color{red!70!black}augensis}. Patients feel {\color{cyan!50!black}they have been victimized by drug abusers since they were 12 years old}. They know {\color{red!70!black}people who don't use marijuana to steal}}\\ \cdashline{1-2}

\model & {\em Whenever there is a basketball or or football, the patient does not really hear what to take. {\color{green!50!black}Patient feels overwhelming} and {\color{cyan!50!black}feels like having this issue since childhood}. Therapist made him realise his problems and patient was curing gradually. Therapist suggested him use. {\color{brown!60!black}The patient was raised Catholic and they feel guilty about spending more money on drugs and parties}. The patient has switched 12 houses in the past year and feel it must have affected their kids. The patient occasionally borrows money and gets less sleep in general}\\ \hline

T5 & \textit{{\color{brown!60!black}The patient is tired when they have to go to work and it's a bit frustrating for them}. They feel tired throughout the day without any food, no panic attacks, no medical condition such as diabetes or stroke. Patients are in a position where they {\color{red!70!black}can} {\color{green!50!black}focus on anything}. They do not want pills to reduce stress hence their life is limited by diet. The patient has lost three pounds in recent weeks due to this fatigue} \\
 \cdashline{1-2}

\model & {\em Whenever the patient goes to work. {\color{green!50!black}The patient is worried that they might have ADHD}. The patient does not suffer from depression, anxiety nor use drugs or call it a metaphysical stuff. The patient wishes to get better and needs something to hold. The patient feel they sway at things, and they have two options. The patient was sent in by a counselor fearing they might hurt themselves. {\color{cyan!50!black}The patient's dad had committed suicide 15 years ago and their sister had attempted once}. {\color{green!50!black}The patient has been diagnosed with depression and anxiety}. {\color{brown!60!black}The patient lives alone}}\\
\bottomrule
\end{tabular}
}
\vspace{-3mm}
\end{table*}

We also show the effect of counseling components and different modules in the pipeline and analyze the results. We demonstrate the importance of our domain-specific modules by evaluating various combinations of counseling components. We show our ablation study in Table \ref{tab:ablation_result}. We observe that masking {SH} and {PD} utterances have a drastic effect on R-1 score as it reports a reduction of over 25 points. Similarly, we observe reductions in other metric as well -- 10 points in R-2, 17 points in R-L, 5 points in QAE, and 33 points in BS. Moreover, we observe similar phenomena in masking any two components (PD and RT; SH and RT) as well. We also perform experiments with masking one component at a time and observe inferior results on average. Overall, we observe that removing the information of any counseling components degrades the performance of \model. Moreover, the importance of CCC module in capturing mental health information is presented in Section \ref{mhic}. Further, we perform ablation with the MH-Know module and observe similar phenomenon with the omission of the MH-Know module as well. 

\subsection{Error Analysis} 
Table \ref{tab:con_sum_features} shows example summaries generated by T5 as well as \model. Sticking to our primary aim of covering mental health-related aspects in summaries, we observe that T5 generates rich semantics; however, lacks in attending critical domain-related phrases in many cases. On the other hand, \model\ encapsulates more domain knowledge and captures critical information inherited from three counseling components. For instance, in the first example in Table \ref{tab:con_sum_features}, T5 covers only the high-level gist of the conversation, viz. \textit{`They secretly talk about incident.'}, whereas \model\ covers essential phrases keeping the semantics intact viz. \textit{`Patient feels overwhelming and feels like having this issue since childhood'}. Moreover, we observe that T5 also commits a few mistakes and generates irrelevant text in both examples (as highlighted in red).

\subsection{Clinical and Human Evaluation}
A team of mental health experts use clinical acceptability framework \cite{Sekhon2017} to evaluate generated summaries. The clinical acceptability framework involves six parameters -- affective attitude, burden, ethicality, coherence, opportunity costs, and perceived effectiveness (c.f. Appendix \ref{app:clinical_val} for more details). Experts rate each summary considering the acceptability parameters on a continuous scale of $0$ to $2$, where higher rating signifies better acceptability. Through expert evaluation, we obtain an average rating of $0.94$ out of $2.00$ -- which falls under the standard acceptability range of $0.70$ to $1.40$ by mental health experts \cite{Sekhon2017}. Hence, the domain experts conclude the summaries to be acceptable for applications in therapy settings. However, in some cases, they pointed out issues with \model\ in comprehending metaphors and subtle transcription errors, which affected the semantic. Tables \ref{tab:rater_eval} and \ref{tab:mental_health_aspects} in Appendix show the acceptability ratings and definition of acceptability parameters, respectively. Overall, {\em mpathic.ai} recommend deploying \model\ with their customers and a commercial pilot with a health-coaching therapy company. 

Furthermore, results of human evaluation on four standard linguistic parameters namely, \textit{relevance} (selection of relevant content), \textit{consistency} (factual alignment between the summary and the source), \textit{fluency} (linguistic quality of each sentence), and \textit{coherence} (structure and organization of summary) are shown in Table \ref{tab:humaneval}. Each parameter is rated on the scale of $1$ to $5$ and found to be qualitatively better than the best baselines. 

\begin{table}[!t]
\centering
\caption{Human evaluation on the summaries generated from \model\ model.}\label{tab:humaneval}
 \vspace{-3mm}
\begin{tabular}{l|cccc}\toprule
Model & Relevance & Consistency & Fluency & Coherence \\
\cmidrule{1-5}
RankAE  & 2.80 & 2.91 & 3.02 & 2.98\\
T5  &2.99& 3.05& 3.04& 2.95\\
\cdashline{1-5}
\textbf{\model} & 3.37 & 3.22 & 3.11 & 3.13\\
\bottomrule
\end{tabular}
\vspace{-5mm}
\end{table}

\section{Mental Health Information Capture (MHIC) metric}
\label{mhic}
\begin{figure}[!t]
  \centering
  \scalebox{0.85}{\includegraphics[width=\columnwidth]{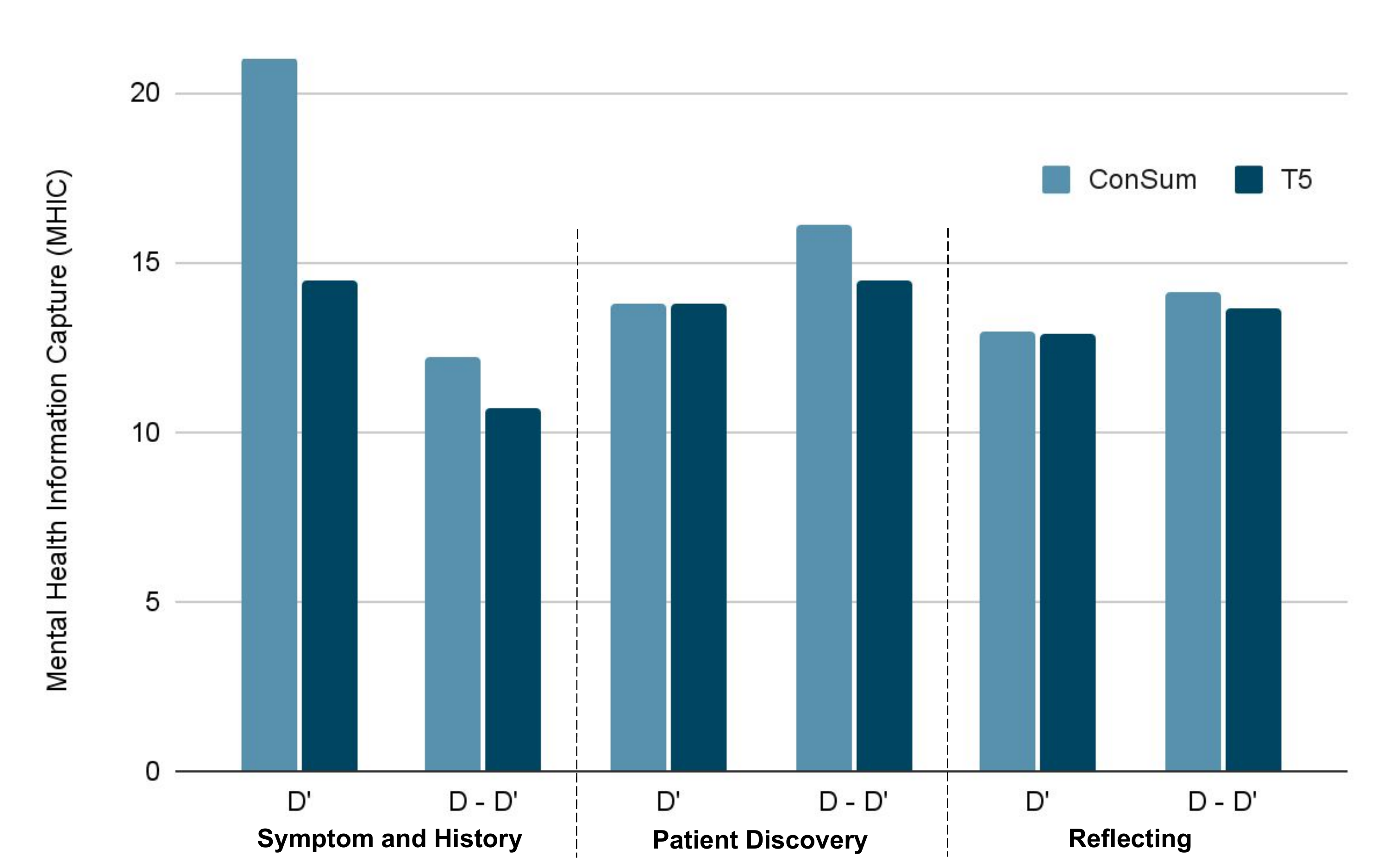}}
  \caption{Comparison between \model\ and T5 summaries and condensed dialogue utterances based on counseling components. The y-axis represents Rouge-1 F1 scores. $D'$ denotes concatenation of utterances tagged with particular counseling component, and $D$ - $D'$ denotes concatenation of utterances excluding corresponding counseling component.}
  \label{fig:subtopics_summary_capture}
  \vspace{-5.9mm}
\end{figure}

Our primary aim is to capture most critical components of mental health counseling leading to successful interventions. Since most of the deep learning methodologies are able to generate contextually and semantically rich text, we observe their incompetence in attaining essential phrases relevant to mental health aspects such as symptom, history of the illness, or other reflecting aspects. Consequently, it becomes extremely challenging to qualitatively assess the summaries and discriminate between the state-of-the-art and domain-specific models. Hence, we propose a new problem-specific metric, MHIC, to mitigate the issue of disregarding such aspects. Mental Health Information Capture (MHIC) attains to the domain-specific counseling components in generated summaries. It incorporates the prediction of counseling components classifier and Rouge 1 score (R-1) to evaluate the generated summaries qualitatively. We calculate MHIC for each counseling component $CL_i \in \{SH, PD, RT\}$ as follows:

$$
    MHIC(CL_i) = {R1\_Score}(S_g,H_i) 
$$

\noindent where $H_i$ represents concatenation of all utterances predicted with $CL_i$ and $S_g$ denotes the generated summary. Figure \ref{fig:subtopics_summary_capture} shows the performance of our model and the best baseline, T5. We also show plots with the inclusion and exclusion of counseling labels for each component. We observe that \model's summaries are more aligned towards the three counseling components. Moreover, we observe that utterances tagged with the SH label have significant representation in generated summaries.

\section{Uniqueness of Mental-health counseling and \model}
Though dialogue summarization in clinical domain \cite{song_summarizing_2020,joshi-etal-2020-dr,inbook} is a well-explored research area, it remains unexplored in the mental health counseling domain. Considering this, in this paper, we raise the research question as ``\textit{\textbf{Why mental health counseling summarization is different than other medical conversation summarization?}}". To answer this question, first we draw significant differences between the two domains and subsequently, show that our proposed model, \model, is specifically designed to handle crucial counseling components of the mental health conversations. Clinical practitioners directly aim for the core problem and the conversation about disease and symptoms are up-front -- usually patients do not shy away discussing their concerns. On the other hand, patients suffering from mental health issues are very much reluctant to speak out their issues in a blunt fashion. 

In our proposed model, we leverage these insights and specifically design various domain-specific sub-modules (DFC, MH-Know, and CCC) to account for such attributes. As a result, \model\ is unique to the mental-health counseling domain instead of other domains.

To demonstrate this, we exploit an existing summarization dataset for clinical conversations in Chinese (ChiCCo) \cite{song_summarizing_2020} and extend it for our purpose. Following \citet{song_summarizing_2020}, we translate the dataset into English using Google Translate. Since \model\ requires counseling psychotherapy elements to operate, we obtain these from the pretrained counseling component and discussion filler classifiers. Subsequently, we train \model\ and other comparative systems on the ChiCCo dataset. 

We show detailed results of the comparative study in Appendix Section \ref{chicco_vs_memo}. In comparison with the encouraging performance of \model\ on \data\ (c.f. Table \ref{tab:baselines}), we argue that \model\ is competent in handling crucial components of the mental health counseling pretty efficiently, as in the absence of such components in other domains, it underperforms.

\section{Commercialization of \model}
\label{sec:evidence}
In this section, we discuss deployment aspects of \model\ as a commercial tool for mental health experts. We lay down two key challenges and resolutions in this respect. 

\noindent \textbf{Challenges:} 
\begin{itemize}[leftmargin=*]
    \item {Ethical concern}: The first critical challenge in the commercialization of \model\ is the usage of publicly available data for training -- we extend HOPE to develop \data. It would be unethical to use it for commercialization. 
    
    \item {Data sparsity}: Even though \model\ performs well on \data\ covering many aspects of mental health components, our experts suggest to include diverse and sophisticated data in the training process, prior to commercialization.
\end{itemize}

\noindent \textbf{Resolutions:} We resolve these challenges in the following way and hope to deploy \model\ as a real-world tool soon. 
\begin{itemize}[leftmargin=*]
    \item We recently signed an MoU with {\em mpathic.ai} to commercially utilize a large-scale conversational dataset to extend the training of \model. 
    
    \item We conduct the experiments and evaluate outputs in two different research labs and an industrial lab at {\em mpathic.ai} to cross-check the consistency and quality of \model.
    
    \item Further, a team of mental health experts from the University of Washington, Wake Forest University, and {\em mpathic.ai} validate the feasibility of \model\ as a commercial application in the real world.
\end{itemize}

\section{Conclusion}
Considering the situation of mental health therapies amid the COVID-19 pandemic, in this paper, we attempted to empower therapists through our research in mental health counseling summarization. To this end, we developed a novel summarization dataset, \data, in a dyadic counseling setup. 
Moreover, to model the counseling summarization in an abstractive fashion, we proposed a domain-enrich transformer architecture, \model. It utilized three domain-centric sub-modules \textit{viz.} a discussion filler classifier, a counseling component classifier, and a PHQ-9 driven mental health knowledge-infused utterance selection module. Our extensive ablation study and the comparative analysis established the superiority of \model\ over several existing models. Furthermore, \model\ is being considered for a large-scale deployment by {\em mpathic.ai}.

\begin{acks}
The authors acknowledge the support of the Ramanujan Fellowship, ihub-Anubhuti-iiitd Foundation set up under the NM-ICPS scheme of the DST, and CAI at IIIT-Delhi. We thank Taichi Shinohara, LICSW, mpathic.ai and Amber Jolley-Paige, PhD, LCPC, mpathic.ai - Wake Forest University - Department of Counseling for providing additional clinical validation.
\end{acks}

\bibliographystyle{ACM-Reference-Format}
\bibliography{sample-base}

\newpage
\appendix

\begin{table*}[t]
\centering
\begin{tabular}{l|cccccc}\toprule
Rater & Affective Attitude & Burden & Ethicality & Intervention Coherence & Opportunity Cost & Perceived Effectiveness\\
\hline
Rater-1  & 0.875 & 1 & 1 & 1.125 & 0.875 & 0.875\\
Rater-2  & 0.75 & 0.875 & 1 & 0.875 & 0.875 & 1\\
\cdashline{1-7}
\textbf{Average} & 0.813 & 0.936 & 1 & 1 & 0.875 & 0.936\\
\bottomrule
\end{tabular}
\caption{Mental health experts rate the summaries generated by \model\  on mental-health aspects.}
\vspace{-5mm}
\label{tab:rater_eval}
\end{table*}

\section{Annotation Process}
\label{annotation_appendix}
Augmenting dialogue utterances with additional information can aid the model to learn the representations better given the limited dataset. The therapists exhibit the three most essential skills, which are -  active listening, observing non-verbal actions, and building rapport \footnote{https://counsellingtutor.com/basic-counselling-skills/}. Elaborating on the techniques followed by therapists in any of their counseling sessions can be categorized as follows: \textbf{Attending} is being in company of a person and giving them your full attention, \textbf{Silence} is to let the patient take complete control of the context, language and pace of their interaction, \textbf{Paraphrasing} lets the patient know that the therapist is actively listening to them by presenting a concise information of context discussed, \textbf{Clarifying questions} makes the therapist understand the patient's issue while avoiding any leading questions which puts the patient at discomfort, \textbf{Focusing} enables the therapist to make decisions with the patient about the issues the patient wants to focus on and disregard other topics, \textbf{Building Rapport} connects the therapist with the patient in an amicable manner, \textbf{Summarizing} is paraphrasing the context discussed till then in a more extended fashion and \textbf{Immediacy} reveals how the therapist feels towards the patient after listening to their events which is regarded as the most necessary skill of a therapist. Finally, \textbf{Homeworks} are given by therapists to patients to follow a particular routine until the next follow-up session, which helps in the patient's path to recovery.\par

To incorporate the information of the techniques discussed above in the dialogue set, we introduce a set of 4 annotation labels considering the structure of therapy counseling sessions carefully and with the help of extensive discussion with leading mental health experts. The annotations are arranged hierarchically in two levels. The first level is binary classification among two labels, namely - \textbf{Counseling components} and \textbf{Discussion Filler}, followed by a sub-classification on counseling components to the following - \textbf{Symptom and History (SH)}, \textbf{Patient Discovery (PD)}, and \textbf{Reflecting (RT)}.\par

\begin{table*}[ht!]
\centering
\begin{tabular}{|c|p{20em}|p{20em}|}\hline
Construct & Definition & Application\\
\hline
Affective attitude & How an individual feels about an intervention & What are your perceptions of the intervention based upon your clinical knowledge\\
\hline
Burden & Perceived amount of effort required to participate & How much effort is required to understand the intervention, consider: spelling, grammar, overall interpretation\\
\hline
Ethicality & Extent to which this is a good fit with your organizations' value system & How does this align with your respective code of ethics? Are there concerns?\\
\hline
Intervention Coherence & Extent to which the intervention is understood & How well the summaries are understood\\
\hline
Opportunity Cost & The extent to which one would benefit from using this intervention & Pros and cons of using this intervention in your respective setting\\
\hline
Perceived Effectiveness & Extent to whcih this intervention will perform in the intended setting & How well this will perform in your clinical setting\\
\hline
\end{tabular}
\caption{Individual metrics as reported in Table \ref{tab:rater_eval} are explained with the application setting.}
\label{tab:mental_health_aspects}
\end{table*}

\section{Implementation Details}
\label{app:implementation}
We used Hugging Face\footnote{https://huggingface.com} to use pre-trained transformers and further fine-tune it for the downstream tasks. All the experiments are carried out on V100 GPUs with GPU RAM of 32GB and 512GB RAM. We discuss the implementation of models below. 
\subsection{Hyperparameters}
\textbf{T5: }We use \emph{t5-base} offered by Simple Tranformers\footnote{https://simpletransformers.ai/} library. For the hyperparameters, namely max\_length, repetition\_penalty, num\_beams, train\_batch\_size, max\_epochs, we set values of 150, 5.0, 8 and 200 respectively.\\
\textbf{SM: }We use the standard framework of this model, which consists of BiLSTM layer for dialogue history encoder and two LSTM networks each for dialogue act labeling and summary decoder. We set only one hidden layer with a size of 256 units for each LSTM network. We experiment using summary attention with a train batch size of 16 and number of epochs as 100.\\
\textbf{PLM: }As followed by the standard pipeline of this model, we use GPT-2 byte pair encoding (bpe) and binarizer provided by Fairseq\footnote{https://github.com/pytorch/fairseq}. We experiment with BART model for the summarizer. The parameters, namely beam\_size, min\_len, max\_len, block\_ngram, are set to 4, 10, 150 and 3, respectively. We train with a batch size of 4 and number of epochs as 200.\\
\textbf{RankAE: } We implement the standard BERT \citep{devlin_bert_2019} encoding for the utterances, which includes the speaker information at the beginning of each utterance. For the encoding, we set shard\_size as 2000, min\_tgt\_tokens as 10 and max\_tgt\_tokens as 150. Following this, the hyperparameters for the main model, namely max\_pos, dec\_dropout, dec\_layers, dec\_hidden\_size, lr\_bert, are set to 512, 0.2, 6, 768 and $1e-3$, respectively. The decoder and encoder follow the same parameters. The train batch size is 4, and number of epochs is 100. \\
\textbf{Pegasus: }From the standard set of pretrained models, we use \emph{pegasus-xsum}. The tokenizer is set to the use the \emph{longest} padding. For the model, we set warmup\_steps, weight\_decay, max\_length as 500, 0.01 and 150, respectively. The train batch size is 4, and num\_train\_epochs is 200, respectively. 

\subsubsection{\model} We first generate embeddings from $distil-bert-uncased$. Further, we implement the model in 3 sub-modular tracks namely, counseling component classifier, discussion filler classifier and MH-Know.  Let's discuss the implementation step by step:\\
\textbf{Counseling component classifier}
The embeddings are passed through a linear layer to bring down the embedding size from $768$ to $100$. We pass the new input to GRU with size set to $100$ and hidden output size is also $100$. Further linear layers bring down the embedding of size $100$ to $4$. We train the sub-module for $20$ epochs with a learning rate of $0.001$ and batch size $4$. We optimize the Adam optimizer using cross-entropy loss.\\
\textbf{MH-Know}
The embeddings are used to calculate the bert score. We use the $0.1.0$ version of bert score to calculate the similarity. The scores obtained are ranged between $0$ to $1$. Here, a bert score of $1$ represents highly similar and $0$ represents highly dissimilar. The summation of all bert score calculated across nine set of lexicons is further used for masking array formation. The mask array is a binary array equivalent to the batch size or dialogue size.\\
\textbf{DFC: }The distil-bert embedding are fine-tuned for binary classification task. On top of encoder, we use a linear layer to condense embedding space from 768 to 100 and further from $100$ to $2$ with a dropout value of $0.20$. We use adam optimizer and cross-entropy loss to perform the training. 

\section{Clinical Validation}
\label{app:clinical_val}
We design an annotation scheme to create a novel dataset in the domain of mental health counseling. The dataset belongs to a critical domain where mental health expert's validation is a must. A team of mental health experts from the University of Washington, Wake Forest University, and {\em mpathic.ai} validated the feasibility of \model\ as an application in the real world. We conducted clinical evaluations from two different experts for clinical acceptability measures. All experts are independently licensed in clinical practice and have backgrounds in automating fidelity of therapy at scale. They included a board-certified, licensed psychologist (SPL) with over 15 years of experience in applying machine learning to psychotherapy evaluation, a licensed clinical social worker specializing in the evaluation of human labeling for machine learning, and doctoral-level counselor and director of clinical AI at {\em mpathic.ai}.

We show the detailed results in Table \ref{tab:rater_eval}. The ratings recorded in Table \ref{tab:rater_eval} are on six different clinical parameters namely, \textit{construct, affective attitude, burden, ethicality, intervention coherence, opportunity cost}, and {\em perceived effectiveness}. The definitions and application use cases of the preceding parameters are mentioned in Table \ref{tab:mental_health_aspects}. 

\section{Counseling vs Clinical Conversation Summarization}
\label{chicco_vs_memo}
In this section we discuss the critical difference between the clinical and counseling conversations. Unlike mental health counseling, clinical conversations are more direct towards the problem. In mental health counseling, it is the job of the therapist to frame the conversation flow and build therapeutic alliance to gain the trust of patients. Another crucial difference is the direction of information (viz issues, symptoms, etc.) flow between the patient and the medical personnel -- in clinical conversations, patients usually convey their issues and symptoms without any prompts from the doctor; whereas, most often than not, the therapist has to nudge and encourage patients to express their issues in counseling sessions. To support this, we conduct a comparative analysis on ChiCCo and MEMO datasets. Table \ref{tab:chicco} shows the results. As expected, \model\ does not perform well due to its highly domain-centric modules. Furthermore, we also observe that T5, which is essentially \model\ without the there domain-specific modules, obtains the overall best performance among all models. 
\begin{table}[t]
\centering
\caption{Results on the clinical conversation ChiCCo dataset. Domain-centric \model\ underperforms in the non-mental health conversation setup. Together with the encouraging performance on \data, it demonstrates the high association of \model\ for the mental health domain.}
\label{tab:chicco}
\vspace{-3mm}
\begin{tabular}{l|rrrrr}\toprule

Model & R-1 & R-2 & R-L & QAE & BS \\
\cmidrule{1-6}

PLM  & $11.82$ & $0.17$ & $14.84$ & $12.37$ & $-1.0890$ \\
RankAE  & $24.00$  & $6.37$ & $20.15$ & $30.25$ & $-0.7466$ \\
SM  & $21.65$ & $7.94$ & $20.37$ & $14.14$ & $-1.001$ \\
Pegasus & $57.54$ & $55.86$ & $27.57$ & $41.10$ & $-0.1253$ \\
T5 &\textbf{71.67} & \textbf{63.62} & \textbf{70.75} & $47.45$ & \textbf{0.2028} \\
\textbf{\model} & $37.13$ & $22.12$ & $17.24$ &\textbf{83.21} & $0.0210$ \\

\bottomrule
\end{tabular}
\vspace{-4mm}
\end{table}


\end{document}